\documentclass{article}

\usepackage{arxiv}

\usepackage[utf8]{inputenc} 
\usepackage[T1]{fontenc}    
\usepackage{hyperref}       
\usepackage{url}            
\usepackage{booktabs}       
\usepackage{amsfonts}       
\usepackage{nicefrac}       
\usepackage{microtype}      
\usepackage{lipsum}
\usepackage{graphicx}
\usepackage{orcidlink}
\usepackage{pgfplots}
\pgfplotsset{compat=1.18}
\usepgfplotslibrary{groupplots}
\usepackage{tikz}
\usetikzlibrary{positioning,arrows.meta,fit,backgrounds,calc}
\usepackage{xcolor}
\usepackage{tabularx}
\usepackage{booktabs}
\usepackage{array}
\newcolumntype{Y}{>{\raggedright\arraybackslash}X}

\definecolor{cProc}{RGB}{31,102,170}
\definecolor{cGen}{RGB}{200,90,30}
\definecolor{cPhys}{RGB}{40,130,90}
\definecolor{cHyb}{RGB}{110,60,150}
\definecolor{cGray}{RGB}{110,110,110}

\graphicspath{ {./images/} }

\title{Curating Synthetic Data for Task-Specific Visual Perception}

\author{
Saptarshi Neil Sinha \\
Fraunhofer IGD, 64283 Darmstadt, Germany
\And
Paul Julius K\"uhn \\
Fraunhofer IGD, 64283 Darmstadt, Germany \And
Michael Weinmann \\
Delft University of Technology, 2628 CD Delft, Netherlands
}

\begin{document}
\maketitle

\begin{abstract}
Synthetic data are most valuable where general-purpose datasets cannot provide
the domain-specific priors a task requires, and where manual annotation is
expensive, imprecise, or infeasible. In this article we argue that the central
question for specialized vision systems is not how to generate more data, but
which data to generate. We therefore discuss \emph{curated} synthetic data,
whose scene content, appearance variations, sensing characteristics, and
annotations are deliberately designed around a given task. We examine three complementary curation
paradigms. Procedural rendering offers explicit control over scene parameters and the annotations follow by
construction. Physically-based simulation encodes the mechanism behind an
observed effect and yields exactly aligned supervision pairs. Generative AI learns sensor-specific appearance from small real seed sets and attains plausible realism, though it remains prone to hallucination and to inaccurate annotation. These paradigms are illustrated with examples from industrial surface defect detection,
restoration of degraded digitized autochrome plates, and 6DoF pose estimation from RGB and
event data. Using these examples, we analyze different data regimes and training strategies
that combine synthetic and real data across these paradigms. We conclude
that curated synthetic data are best understood as a complement to real
observations, and that hybrid pipelines combining controllable supervision with
learned appearance are the most promising direction for reliable sim-to-real
transfer. 
\end{abstract}
Synthetic data has become an important alternative to manually collected and annotated data in computer vision.
%
Real-world datasets can be expensive to acquire, difficult to annotate, and inherently limited in their coverage of rare defects, unusual conditions, rare objects, or specialized sensor modalities. These limitations become particularly pronounced in task-specific applications, where even a large general-purpose dataset may not contain the appearance variations, sensing characteristics, or annotations required by the target system.
%
\begin{figure*}
    \centering
    \includegraphics[width=\textwidth]{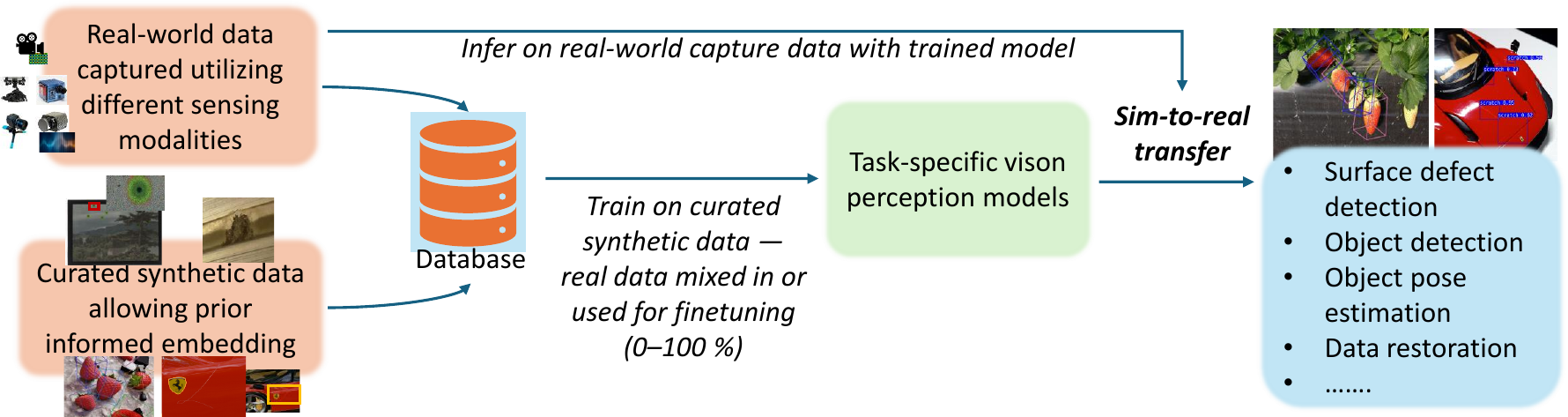}
    \caption{Workflow for curated synthetic data in task specific multimodal
    scene understanding. Real world data from different sensing modalities is
    combined with curated synthetic data that embeds task specific prior
    knowledge. Models are trained using synthetic and real data, with the real
    data share varying from 0 to 100\% or used for fine tuning. The trained
    models are then transferred to real applications through sim to real
    transfer, including surface defect detection and object pose estimation.}
    \label{fig:concept}
\end{figure*}
Synthetic data address this problem by generating images or sensor measurements from a controllable description of the scene, while providing high-quality annotations as a direct consequence of the generation process, thereby mitigating costly manual annotation processes. Previous investigations have therefore leveraged synthetic data for a wide range of computer vision tasks, including human pose estimation~\cite{shotton:2011}, optical flow, stereo, and depth estimation~\cite{Mayer_2016_CVPR}, semantic segmentation~\cite{richter:2016}, object detection and tracking~\cite{detection_tracking}, material recognition under varying, uncontrolled illumination conditions~\cite{material_classification}, reflectance estimation~\cite{reflectance_estimation}, 3D reconstruction~\cite{3d_reconstruction} and inverse rendering~\cite{inverse_rendering}, thereby demonstrating great potential for autonomous driving~\cite{survey_autonomous_driving} and intelligent manufacturing~\cite{shafiee:2026}.
%
These examples show that synthetic data transfers to real applications whenever it captures the
variations relevant to the target task.
%
%
%

%
In this paper, we therefore focus on curated synthetic data, whose scene
content, appearance, and annotations are deliberately designed around a
particular task, object, and sensor. This task-specific curation allows domain knowledge to be embedded directly into the generation process. Geometry, materials, illumination, camera parameters, defect characteristics, degradation processes, object motion, and sensor response can be controlled according to the requirements of the application. At the same time, the parameters used to generate the data provide annotations without the cost of manual labeling. To examine how synthetic data can be curated rather than merely generated at scale, we focus on task-specific priors, the trade-off between controllability and realism, and transfer to real sensors and deployment platforms.
%
Figure~\ref{fig:concept} illustrates this workflow.
%
Real-world observations from different sensing modalities provide information about the target domain, while curated synthetic data allow task-specific priors to be embedded systematically, e.g., through material models, procedural defect descriptions, effect simulation and/or simulated camera motion. Models can then be trained using synthetic and real data in different proportions, or pretrained on synthetic data and subsequently fine-tuned using real observations. The objective is not necessarily to replace real data, but to use synthetic data to compensate for missing or expensive examples and to improve the efficiency of the available annotation budget.
%

%
We examine task-specific synthetic data across three application settings, progressing from 2D visual defects to 3D object understanding and alternative sensing modalities.
%
First, we study industrial surface
defect detection using procedural rendering and generative AI. Procedural
rendering provides explicit control over scene parameters and
annotations, whereas diffusion models can reproduce defect appearance and
sensor characteristics learned from a small set of real images. Second, we
consider image restoration of degraded autochrome plates. We simulate the physical processes of the degradation to generate aligned clean and degraded image pairs, enabling training
when corresponding real pairs are unavailable. Third, we examine 6DoF pose
estimation using synthetic RGB and event data. At the example of 6D pose estimation for strawberries, we demonstrate the effect of realistic, distractor-augmented synthesized training data for accurate pose estimation and the deployment on both a high-performance GPU, the
NVIDIA RTX 3090, and an edge device, the NVIDIA Jetson Orin Nano. Furthermore, we show  how synthetic pose data can be extended to event cameras and applications involving fast motion or limited
computational resources at the example of the The YCB-Ev SD dataset~\cite{eventposeestimation}.
%
%
These applications illustrate that synthetic-data generation must be tailored to the task: defect detection requires appropriate variation in defect morphology and appearance, restoration requires precise control of degradation, and pose estimation requires accurate geometry, pose, and sensing characteristics. They also highlight a trade-off between procedural and generative approaches. Procedural methods provide control and reliable ground truth but may miss real-world appearance and sensor statistics, whereas generative methods can yield more plausible and diverse defects once adapted to the target domain, but are prone to hallucinations and reduced annotation accuracy. Hybrid approaches can therefore combine the strengths of both. Across the presented applications, curated synthetic data serve primarily as a complement to real observations, providing targeted variation and reliable supervision where real data are scarce, difficult to annotate, or impractical to collect.
\begin{figure*}[t]
\centering
\includegraphics[width=\textwidth]{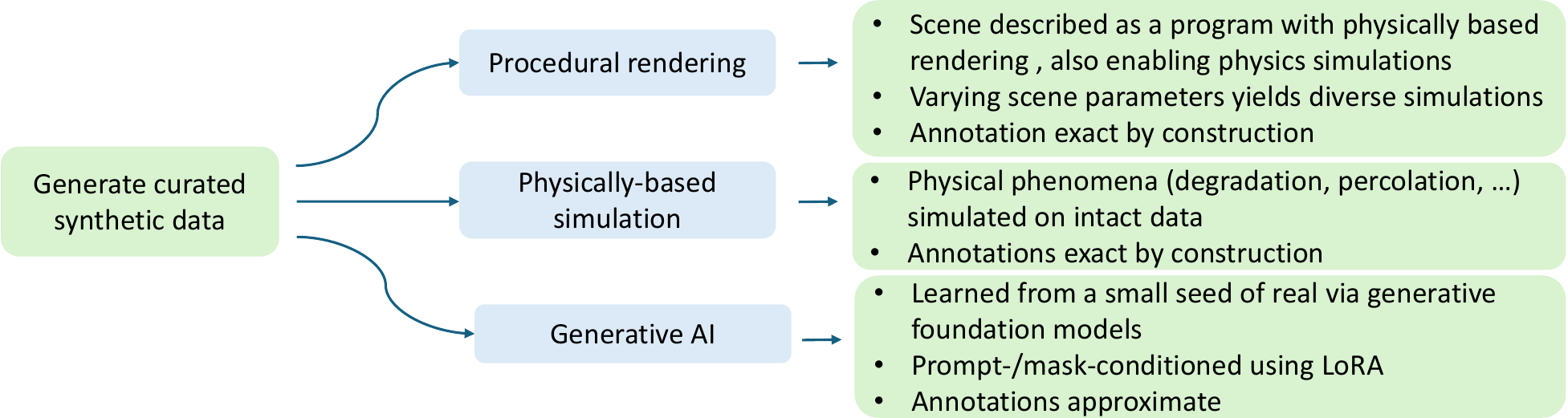}
\caption{\textbf{Curation of synthetic data utilizing various approaches.} Curation embeds
task-specific priors through three complementary simulation paradigms.
\emph{Procedural rendering} and \emph{physically-based simulation}
yield annotations that are exact by construction but reproduce only
explicitly modelled variations, whereas \emph{simulation utilizing generative AI}
learns realistic texture, contrast and sensor statistics from a small real
seed set at the cost of annotation accuracy.}
\label{fig:taxonomy}
\end{figure*}
\section{Curation of synthetic data}
\label{sec:curation}
In this section we discuss three main approaches to curating synthetic data: procedural rendering, physically-based simulation, and generative AI approaches (see Fig.~\ref{fig:taxonomy}). In the following, we briefly discuss these approaches and also provide a discussion of resulting trade-offs.
\noindent\textbf{Procedural rendering}
The generation of training data via procedural rendering approaches~\cite{denninger2023blenderproc2,spear} is built on the concept of defining a scene as a program rather than as a fixed asset. Scene parameters such as geometry, materials, lighting, etc. are expressed as parameterized rules, and each sample is produced by drawing those parameters and rendering the result with a physically-based renderer. Two properties make this paradigm attractive for curation, i.e.\ the sampled configuration space is
not limited to what real acquisition happened to capture, and every scene parameter is known at
render time, so masks, boxes, depth and metric 6DoF poses come for free. This is decisive wherever
annotations cannot be produced manually at competitive accuracy, which makes procedural rendering the
natural choice for 3D scene understanding and, since the camera trajectory is part of the scene
program, also for temporal modalities such as event streams, at the price of a rendering budget that
turns compute into the dominant constraint. 
Procedural rendering has two main limitations, i.e.\ it requires explicit assets in the form of CAD data, a scanned model or geometry modeled by a designer, together with a controllable material description, and it reproduces only explicitly modeled variations, so that the texture, contrast and sensor statistics of a specific acquisition setup are approximated rather than reproduced. Throughput is the second constraint, since
pipelines such as BlenderProc2~\cite{denninger2023blenderproc2} offer no built-in distributed
execution and scale poorly to large synthetic sets, which can be partially mitigated by real-time
engine-based simulators such as SPEAR~\cite{spear} that render photorealistic images together with
ground-truth modalities at interactive frame rates under batched, programmatic scene control.
\begin{figure*}[t]
\centering
\begin{tikzpicture}
\begin{groupplot}[
  group style={group size=2 by 1, horizontal sep=1.5cm},
  width=0.4\textwidth, height=4.8cm,
  xlabel={Fraction of real training data (\%)},
  xtick={10,25,50,100},
  xmin=2, xmax=112,
  grid=major, grid style={gray!25, densely dotted},
  tick label style={font=\small},
  label style={font=\small},
  title style={font=\small\bfseries, yshift=1pt},
  every axis plot/.append style={line width=1.1pt, mark size=2.6pt},
  legend style={font=\small, draw=none, legend columns=-1,
    /tikz/every even column/.append style={column sep=10pt}},
  legend to name={scalinglegend},
  clip=false,
]
\nextgroupplot[
  title={(a) YOLO26 (CNN)}, ylabel={mAP@50 on real test set},
  ymin=0.02, ymax=0.78, ytick={0.1,0.2,0.3,0.4,0.5,0.6,0.7},
  yticklabel style={/pgf/number format/fixed,
                    /pgf/number format/precision=1},
]
\addplot[dashed, gray, forget plot]
  coordinates {(2,0.4513) (112,0.4513)};
\node[font=\scriptsize, gray!70!black, anchor=south west, inner sep=1pt]
  at (axis cs:60,0.4513) {synthetic only};
\addplot[dashdotted, red!60!black, forget plot]
  coordinates {(2,0.6100) (112,0.6100)};
\node[font=\scriptsize, red!60!black, anchor=south west, inner sep=1pt]
  at (axis cs:57,0.6100) {real only, 100\%};
\addplot[mark=square*, color=red!75!black,
  error bars/.cd, y dir=both, y explicit]
 coordinates {(10,0.0561) +- (0,0.0076)
              (25,0.0428) +- (0,0.0053)
              (50,0.3229) +- (0,0.0364)
              (100,0.6100)+- (0,0.0255)};
\addlegendentry{real only}
\addplot[mark=*, color=blue!70!black,
  error bars/.cd, y dir=both, y explicit]
 coordinates {(10,0.4319) +- (0,0.0115)
              (25,0.5979) +- (0,0.0259)
              (50,0.6722) +- (0,0.0023)
              (100,0.7064)+- (0,0.0083)};
\addlegendentry{synthetic (WB) $+$ real}
\addplot[only marks, mark=star, mark size=5pt, color=green!45!black]
 coordinates {(100,0.7227)};
\addlegendentry{fine-tuned from synthetic}
\draw[gray!60!black, line width=0.5pt]
  (axis cs:100,0.7064) -- (axis cs:88,0.746);
\node[font=\scriptsize, anchor=east, gray!60!black]
  at (axis cs:88,0.748) {$+0.016$};
\nextgroupplot[
  title={(b) LW-DETR (transformer)},
  ylabel={mAP@50 on real test set},
  ymin=0.18, ymax=0.78, ytick={0.2,0.3,0.4,0.5,0.6,0.7},
  yticklabel style={/pgf/number format/fixed,
                    /pgf/number format/precision=1},
]
\addplot[dashed, gray, forget plot]
  coordinates {(2,0.3817) (112,0.3817)};
\node[font=\scriptsize, gray!70!black, anchor=south west, inner sep=1pt]
  at (axis cs:60,0.3817) {synthetic only};
\addplot[dashdotted, red!60!black, forget plot]
  coordinates {(2,0.5350) (112,0.5350)};
\node[font=\scriptsize, red!60!black, anchor=south west, inner sep=1pt]
  at (axis cs:57,0.5350) {real only, 100\%};
\addplot[mark=square*, color=red!75!black, forget plot,
  error bars/.cd, y dir=both, y explicit]
 coordinates {(10,0.2380) +- (0,0.0234)
              (25,0.3553) +- (0,0.0272)
              (50,0.4070) +- (0,0.0397)
              (100,0.5350)+- (0,0.0303)};
\addplot[mark=*, color=blue!70!black, forget plot,
  error bars/.cd, y dir=both, y explicit]
 coordinates {(10,0.3867) +- (0,0.0187)
              (25,0.5113) +- (0,0.0327)
              (50,0.6357) +- (0,0.0246)
              (100,0.6830)+- (0,0.0221)};
\addplot[only marks, mark=star, mark size=5pt,
         color=green!45!black, forget plot]
 coordinates {(100,0.6910)};
\draw[gray!60!black, line width=0.5pt]
  (axis cs:100,0.6830) -- (axis cs:88,0.726);
\node[font=\scriptsize, anchor=east, gray!60!black]
  at (axis cs:88,0.728) {$+0.008$};
\end{groupplot}
\end{tikzpicture}
\caption{\textbf{Effect of adding real images to procedurally rendered defect data.} Detection performance
(mAP@50) on the \emph{real} toy-Ferrari test split~\cite{scratchsim} for increasing fractions of real
training data. Real-only training (red squares) collapses below a real share of $50\,\%$, whereas
mixing in curated synthetic renderings (blue circles) sustains high performance with only $10\,\%$
real images and already exceeds the full real-only baseline (dash-dotted) at $50\,\%$. Fine-tuning
from synthetic weights (green $\star$) is best overall, but its margin over mixed training at
$100\,\%$ real data is small ($+0.016$ YOLO26, $+0.008$ LW-DETR).}
\label{fig:scaling}
\end{figure*}
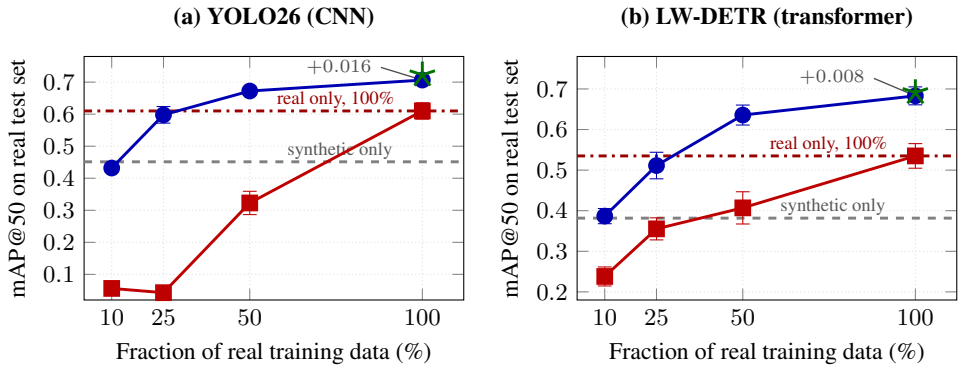
\begin{table}[t]
\centering
\caption{Curation paradigms and their properties.}
\label{tab:paradigms}
\scriptsize
\setlength{\tabcolsep}{6pt}
\renewcommand{\arraystretch}{1.05}
\scalebox{1}{
\begin{tabularx}{\columnwidth}{@{}
  >{\raggedright\arraybackslash}p{0.15\columnwidth}
  >{\raggedright\arraybackslash}X
  >{\raggedright\arraybackslash}X
  >{\raggedright\arraybackslash}X@{}}
\toprule
 & Procedural rendering & Physical process & Generative AI \\
\midrule
Specification & Scene program & Process model & Learned from seeds \\
Annotation & Exact & Exact & Approximate \\
Realism & Asset-limited & Inherited from input & Learned from domain \\
Diversity & Sampled parameters & Model expressiveness & Seed set \\
Limitation & Throughput & Model design & Annotation quality, hallucinations \\
\bottomrule
\end{tabularx}}
\end{table}
\noindent\textbf{Physically-based simulation}
Data generation based on physically-based simulation is focused on integrating priors regarding the processes acting upon the scene, such as degradation, percolation or other physics-based effects. 
The effect of interest, for example a degradation of color, is formulated
analytically 
based on a mathematical formulation of the underlying spatio-temporal degradation effect, modeling its spatial extent, its color response and its interaction with the respective materials in the scene. In turn, this model is then applied on intact scene observations to augment the data with respective effect characteristics.
Since the effect is applied to a known input, the clean and the affected observation are
perfectly aligned by construction and the affected region is exact by definition, which makes
this the only paradigm that provides such pairs when the undegraded counterpart of a real
observation does not exist, while the realism of geometry, texture and sensor characteristics
is inherited from the real input. The main control parameter is the process model itself, whose parameterization determines the
appearance of the effect and hence also its limitation, since the achievable diversity is bounded by
the expressiveness of the formulation and effects not described by the model do are not captured in the data.
\noindent\textbf{Simulation utilizing generative AI}
A further approach is data generation via a generative AI model.
Instead of
modeling geometry, material, illumination or process analytically, a generative model is
adapted to a small annotated seed set of real observations and thereby captures the sensor statistics of the  acquisition setup. Since these are exactly the
properties that are hard to specify by hand, generated samples can appear more realistic than
rendered ones and transfer better to the real task. Control is exercised through conditioning
and filtering rather than through scene parameters. Three mechanisms provide the necessary control, i.e.\ domain adaptation of the generative model, for
example with low-rank adaptation~\cite{hu2021loralowrankadaptationlarge}, to align the generated
appearance with the target domain, mask-guided inpainting into unaffected real observations to
constrain where an effect appears, and a filtering stage to remove degenerate samples. The main
limitation is lack of accurate annotation, since the spatial support of the generated effect is unknown and must be
recovered with a promptable segmentation model, which is reliable for compact, well-contrasted
structures but not for thin, low-contrast ones, so annotation accuracy rather than image quality
bounds this paradigm. 
\noindent\textbf{Trade-offs and hybrid curation}
%
In Table~\ref{tab:paradigms} we present a summary of the resulting trade-off. The model-based paradigms provide exact annotations, yet their validity extends only as far as the modeled characteristics, since they reproduce only explicitly modeled variations. The generative paradigm, in contrast, matches the target appearance once adapted to the target domain, at the cost of approximate annotations and potential hallucinations. Both observations suggest a hybrid curation in which a model-based stage provides reliable ground truth and a generative stage aligns the appearance with the target sensor.
\section{Applications}
In the following, we provide exemplary applications for the different curation paradigms in
Figure~\ref{fig:taxonomy}, progressing from 2D defect localization to restoration, and
further to 3D pose estimation and an alternative sensing modality. 
\begin{figure*}[!htbp]
\centering
\begin{tikzpicture}
\begin{groupplot}[
  group style={group size=2 by 1, horizontal sep=2cm},
  width=0.4\textwidth, height=4.8cm,
  grid=major, grid style={gray!25, densely dotted},
  tick label style={font=\scriptsize},
  label style={font=\scriptsize},
  title style={font=\scriptsize\bfseries, yshift=1pt},
  every axis plot/.append style={line width=1.0pt, mark size=2.3pt},
  xlabel={Real share of training budget (\% of full real split)},
  xtick={0,25,50,75,100},
  xmin=-6, xmax=106,
  yticklabel style={/pgf/number format/fixed,
                    /pgf/number format/precision=2},
]
\nextgroupplot[
  title={(a) BSData (pitting)},
  ylabel={AP$_{50}$},
  ymin=0.70, ymax=0.99,
  ytick={0.70,0.75,0.80,0.85,0.90,0.95},
  legend style={font=\tiny, draw=none, at={(0.98,0.03)},
                anchor=south east, fill=white, fill opacity=0.85,
                text opacity=1, row sep=0.5pt},
]
\addplot[blue!55!white,   densely dashed, line width=0.7pt, forget plot]
  coordinates {(-6,0.933) (106,0.933)};
\addplot[violet!55!white, densely dashed, line width=0.7pt, forget plot]
  coordinates {(-6,0.951) (106,0.951)};
\addplot[orange!60!white, densely dashed, line width=0.7pt, forget plot]
  coordinates {(-6,0.882) (106,0.882)};
\addplot[mark=*, color=blue!70!black]
  coordinates {(0,0.729) (25,0.871) (50,0.906) (75,0.926) (100,0.926)};
\addlegendentry{YOLOv26}
\addplot[mark=square*, color=violet!70!black]
  coordinates {(0,0.775) (25,0.880) (50,0.912) (75,0.909) (100,0.924)};
\addlegendentry{LW-DETR}
\addplot[mark=triangle*, mark size=2.9pt, color=orange!85!black]
  coordinates {(0,0.730) (25,0.825) (50,0.841) (75,0.855) (100,0.865)};
\addlegendentry{YOLOX-S}
\node[font=\tiny, gray!60!black, anchor=north west]
  at (axis cs:-4,0.987) {dashed: union (100\,\%\,R $+$ synthetic)};
\nextgroupplot[
  title={(b) MSD (scratch)},
  ylabel={AP$_{50}$},
  ymin=0.74, ymax=1.02,
  ytick={0.75,0.80,0.85,0.90,0.95,1.00},
  legend style={font=\tiny, draw=none, at={(0.98,0.03)},
                anchor=south east, fill=white, fill opacity=0.85,
                text opacity=1, row sep=0.5pt},
]
\addplot[teal!55!white, densely dashed, line width=0.7pt, forget plot]
  coordinates {(-6,0.990) (106,0.990)};
\addplot[mark=diamond*, mark size=2.9pt, color=teal!70!black]
  coordinates {(0,0.770) (25,0.970) (50,0.984) (75,0.990) (100,0.995)};
\addlegendentry{YOLOv26}
\node[font=\tiny, gray!60!black, anchor=north west]
  at (axis cs:-4,1.015) {dashed: union (100\,\%\,R $+$ 100\,\%\,S)};
\end{groupplot}
\end{tikzpicture}
\caption{\textbf{Effect of adding real images to generatively synthesized defect data.} Solid curves report
AP$_{50}$ on the real test split as the real share of a fixed-size training set grows from
$0\,\%$ (synthetic only) to $100\,\%$ (real only), while dashed lines mark the best union setting, in
which synthetic images are \emph{added} on top of the complete real set.
\textbf{(Left)} On BSData (pitting), AP$_{50}$ saturates beyond a real share of $50\,\%$, so
synthetic images can replace about half of the annotation budget at a cost of
$\leq 0.02$~AP$_{50}$, and all union lines exceed the real-only endpoint ($+0.007$ YOLOv26,
$+0.027$ LW-DETR, $+0.017$ YOLOX-S).
\textbf{(Right)} On the near-saturated MSD scratch subset ($0.995$ real only) the substitution trend
holds, but the union line drops slightly below the baseline ($-0.005$), indicating that generated
data helps mainly where the real task is not already solved.}
\label{fig:synsur_map50}
\end{figure*}
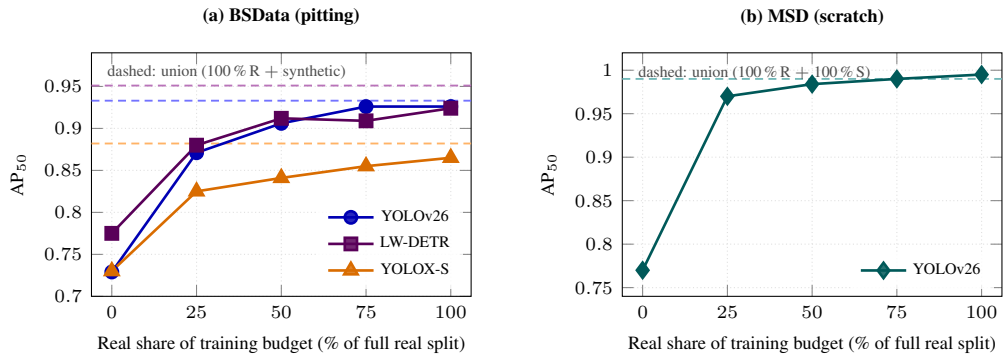
\noindent\textbf{Industrial surface defect detection}
Industrial surface inspection is a prime example of a task where data, not model capacity, is
the bottleneck. Defects occur on a small fraction of produced parts, annotation requires
expert knowledge, and the resulting datasets are small and strongly imbalanced. The two
studies below address this setting with different curation paradigms, depending on whether
explicit assets or a small annotated seed set are available.
\noindent\textbf{\textit{Procedural rendering}}
ScratchSim~\cite{scratchsim}, a procedural scratch
generation pipeline built on BlenderProc2~\cite{denninger2023blenderproc2}, is a recent example of leveraging data curation based on procedural rendering in the context of surface defect generation. This approach combines a CAD
model of the inspected part with procedurally generated scratch maps, a physically-motivated
material model and randomized surroundings, so that the defect annotations follow directly from
the sampled scratch parameters. 
The pipeline was applied to two objects with deliberately different surface
characteristics, a matte powder-coated industrial grip and a glossy toy Ferrari car,
so that curation choices can be varied while the detection task and the real test set
stay fixed. Two small detectors were considered, i.e., a convolutional one (YOLO26
Nano~\cite{jocher2026ultralyticsyolo26unifiedrealtime}) and a transformer-based one
(LW-DETR Tiny~\cite{chen2024lw}), reflecting deployment on edge devices. As shown in the work by Sinha et~al.~\cite{scratchsim} and in Figure~\ref{fig:scaling},
detection performance of real-only training collapses in the scarce-data regime, whereas
adding synthetic renderings removes this collapse and already surpasses the full real-only
baseline at half the real annotation budget. How the two domains are combined matters
comparatively little, i.e., fine-tuning from synthetic weights is best overall but only
marginally ahead of mixing, while synthetic-only training stays clearly below the real
baseline. The matte industrial grip with YOLOX~\cite{yolox2021,scratchsim} reproduces the
same trend, so the behavior is not tied to a specific object, material or detector family.
\begin{figure*}
    \centering
    \includegraphics[width=0.9\linewidth]{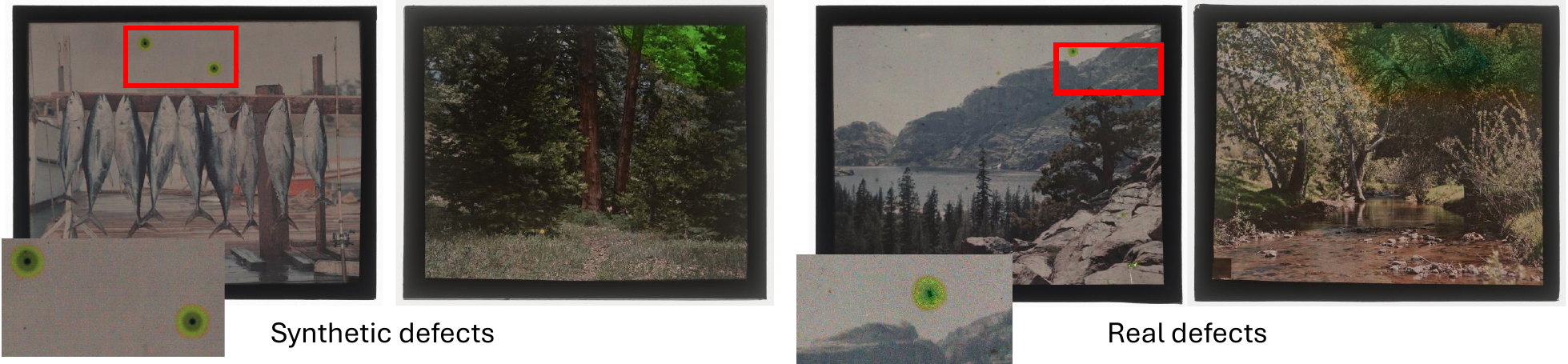}
    \caption{Synthetic autochrome greening defects generated using a signal processing approach~\cite{autochromesgreening}}
    \label{fig:autochrome_signal}
\end{figure*}
\begin{figure}
    \centering
    \includegraphics[width=0.4\linewidth]{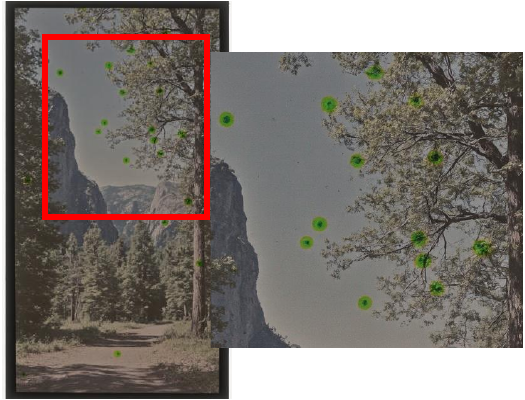}
    \caption{Autochrome greening defects generated utilizing generative AI~\cite{text2Autochromes_dh25}}
    \label{fig:autochrome_generative}
\end{figure}
\noindent\textbf{\textit{Simulation utilizing generative AI}}
When neither a CAD model nor a controllable material description but only a small
annotated seed set of real defects is available, simulations based on generative AI approaches still offer an alternative to synthesize respective training data to train defect detection models. An exemplary work is SynSur~\cite{synsur}, which derives a defect prompt
from a vision-language model, adapts a text-to-image model with
LoRA~\cite{hu2021loralowrankadaptationlarge}, inserts defects into defect-free images by
mask-guided inpainting, filters the candidate pool and derives the labels automatically,
producing 420 annotated images in roughly eight GPU hours. The evaluation on
BSData~\cite{synsur} with small low-contrast pitting defects and the scratch subset of
MSD~\cite{synsur} with thin screen scratches from Kühn et~al.~\cite{synsur} and summarized in Figure~\ref{fig:synsur_map50} shows an assessment of how far synthetic images can substitute for annotated real ones. For this purpose, 
the share of real images is varied within a
training set of \emph{fixed size}. Performance grows with the real share and saturates beyond
$50\,\%$ on BSData, so about half of the annotation budget can be shifted to synthetic images at a
negligible cost in accuracy. Adding synthetic images on top of the complete real set improves all
three detectors moderately on BSData but slightly degrades the already saturated scratch baseline,
so synthetic data helps where the real task is not yet solved. The ablations identify domain
adaptation as essential, since prompt engineering alone yields visually plausible but
domain-inconsistent defects, while candidate filtering by perceptual similarity and prompt
alignment corrects the strong bias toward very small defects in the raw pool. The largest
bottleneck was annotation quality, as a promptable segmentation model refines the generated masks
reliably for pitting defects but fails on thin low-contrast scratches, where the inpainting mask
itself has to serve as a conservative surrogate label.
\noindent\textbf{Restoration of degraded autochromes}
To assess the simulation of physical processes as a curation paradigm, we exemplarily select the restoration of
digitized autochrome plates, the first commercially successful colour photography process, which
suffer from a characteristic degradation known as greening~\cite{autochromesgreening} resulting from  the bleeding of the unstable green dye out into the surrounding regions under environmental factors such as humidity, light, etc. Since usually the
clean version of an original plate is not available anymore, supervised restoration cannot be trained on
pairs of real images and degraded images. Therefore, physically-based simulation of respective greening degradation,
taking inspiration from how the green dyes bleed, can be applied to intact plate regions 
(Fig.~\ref{fig:autochrome_signal}), which yields perfectly aligned clean and degraded image pairs at
scale and allows the restoration network to be trained on purely synthetic data~\cite{autochromesgreening}. As an alternative approach, such degradations can instead be produced with generative
AI techniques~\cite{text2Autochromes_dh25}, which yields visually more plausible defects
(Fig.~\ref{fig:autochrome_generative}) at the cost of far weaker control over their exact spatial
extent, the same failure mode observed for thin, low-contrast scratches in SynSur. Since the
affected region serves as the supervision target itself, the analytical formulation is preferable
here. The resulting restorations are plausible and compare favourably to state-of-the-art editing tools
such as Adobe Photoshop Generative Fill, which alter global colour and texture
statistics~\cite{autochromesgreening}, whereas the intervention stays confined to the affected
region, an essential property for heritage applications. The output still needs to be refined and
colour graded by a conservator and should be presented alongside the original plate so that every
intervention remains traceable. Within these constraints the main benefit is assisting conservators
in reaching the same result much faster, with the required time reduced to roughly one quarter for
larger defective areas in an expert-based analysis.
\begin{figure*}[htb!]
    \centering
    \includegraphics[width=0.9\linewidth]{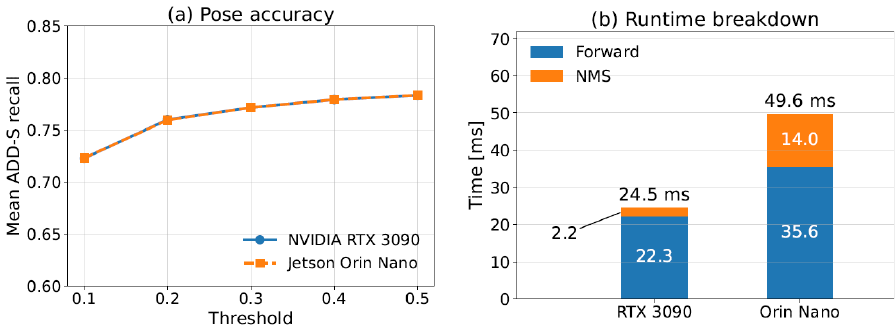}
    \caption{Evaluation of the YOLOX 6D Pose model trained exclusively on
    synthetic data. (a) Fraction of test instances whose ADD S error is below
    the specified threshold, averaged over classes. (b) Inference time divided
    into the forward pass and non maximum suppression for the NVIDIA RTX 3090
    and the Jetson Orin Nano.}
    \label{fig:pose_accuracy_runtime}
\end{figure*}
\begin{figure}
    \centering
    \includegraphics[width=0.8\linewidth]{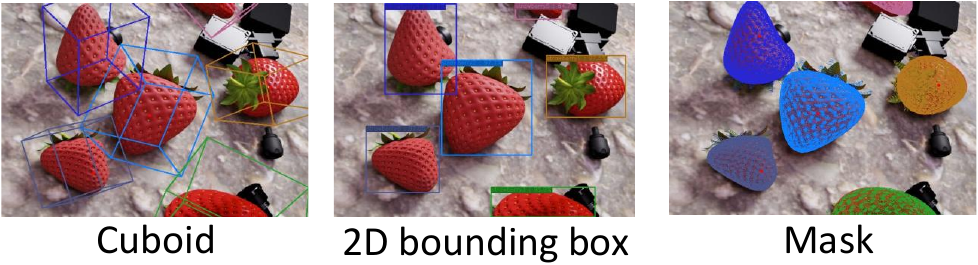}
    \caption{\textbf{Synthetic data for 6DoF strawberry pose estimation.} Rendered RGB image with instance mask, 3D cuboid, and 2D bounding box annotations~\cite{6dstrawberry}.}
    \label{fig:strawberry_synthetic_data}
\end{figure}
\noindent\textbf{6DoF pose estimation}
When lifting the task from 2D to 3D, the annotation itself plays the critical role, since a
metrically exact 6DoF pose cannot be obtained by manual labeling of real images. Therefore, we can assess the potential of
procedural rendering as a curation paradigm as exemplarily shown on the following two applications that share the same detector but differ in
the sensing modality.
\noindent\textbf{\textit{Selective harvesting from rendered RGB data}}
Selective harvesting requires a robotic arm to approach and separate a fruit in a cluttered
environment, which a 2D detection combined with a depth measurement does not provide. In the case
6DoF strawberry pose estimation~\cite{6dstrawberry}, training data can be generated from a set of digitized strawberry models
varying in shape, size, hue and subsurface scattering. These models can then be placed in scenes using
BlenderProc2, where physics-based settling creates realistic placement,
YCB-Video distractors~\cite{xiang2018posecnn} introduce occlusions and
CCTextures~\cite{denninger2023blenderproc2} provide the backgrounds. This pipeline yields RGB
images, depth maps, instance masks and BOP-format pose annotations
(Fig.~\ref{fig:strawberry_synthetic_data}), and YOLOX 6D Pose~\cite{yolo6d_pose} is trained
exclusively on this data. As seen in Fig.~\ref{fig:pose_accuracy_runtime}(a), the purely synthetically trained model achieves
virtually identical ADD-S accuracy on an NVIDIA RTX 3090 and a Jetson Orin Nano across all
thresholds. These devices differ only in inference speed ($24.51$~ms versus $49.59$~ms), a gap
dominated by non-maximum suppression (Fig.~\ref{fig:pose_accuracy_runtime}(b)) that still leaves the
edge device fast enough for power-constrained robotic systems.
\noindent\textbf{\textit{Synthetic event data}}
This paradigm also extends to other sensing modalities, which we assess at the example of the
YCB-Ev SD dataset~\cite{eventposeestimation}, which renders physically-based
scenes with BlenderProc2~\cite{denninger2023blenderproc2} and converts short
high-frame-rate sequences under linear camera motion into event streams.
Generating the dataset required millions of rendered frames and about one week
on six GPUs, indicating the need for distributed rendering when generating large datasets. Moreover, event data
has no single standard representation, which turns the encoding into an
additional curation decision, i.e., training the same YOLOX-6D-Pose model on
different encodings shows that polarity is the most important cue and that a
time-surface encoding more than triples the ADD(-S)~$0.5d$ score compared with a
plain event histogram. Curation therefore does not end at the rendered scene,
and validation on real event recordings remains an important direction for
future work. Nevertheless, the potential of such data for 6DoF object pose estimation~\cite{eventposeestimation} is encouraging, since this modality is
particularly suited to fast-moving objects such as UAVs~\cite{magrini2025fred}
and to adverse weather or low-light conditions, and synthetic data can extend it
to previously unseen objects simply by varying the underlying CAD models.

\section{Conclusion}
Synthetic data are most effective when they are curated for the specific failure
modes, annotation requirements, objects and sensors relevant for a particular task. They help where real
data are scarce, provide accurate labels for problems that cannot be annotated
manually, and enable experiments with difficult sensing modalities. At the same
time, procedural rendering and physically-based simulation reproduce only the
variations that are modeled explicitly, whereas generative AI, once adapted to target domain, yields more plausible
appearance but can introduce annotation errors, as confirmed by the aforementioned defect detection,
autochrome restoration and pose estimation applications. Curated synthetic
data therefore complement real observations rather than replace them, and gains
shrink once the real task is already solved. Future work should focus on combining
procedural rendering and physically-based simulation for reliable geometry and
supervision with generative AI for realistic textures, lighting and sensor
characteristics, a promising path toward reliable sim-to-real transfer that
preserves both accurate annotations and the appearance as captured by the sensor.

\bibliographystyle{unsrt}
\bibliography{synthetic}

\end{document}